\documentclass[letterpaper, 10 pt, conference]{ieeeconf}  

\IEEEoverridecommandlockouts                              

\usepackage{mathtools}
\usepackage{subcaption}
\usepackage{cancel}

\title{\LARGE \bf A Passivity-based Concurrent Whole-Body Control (cWBC) of\\Persistently Interacting Human-Exoskeleton Systems}

\author{Federico L. Moro$^{1*}$, Niccol\`o Iannacci$^{1,2}$, Giovanni Legnani$^3$, and Lorenzo Molinari Tosatti$^1$
\thanks{$^{1}$Institute of Industrial Technologies and Automation (ITIA), National Research Council (CNR) of Italy, Via Corti 12, 20133 Milano, Italy}%
\thanks{$^{2}$Department of Mechanical Engineering, Politecnico di Milano, Via Giuseppe La Masa 1, 20156, Milano, Italy}%
\thanks{$^{3}$Department of Mechanical and Industrial Engineering, Universit\'a degli Studi di Brescia, Via Branze 38, 25123, Brescia, Italy}%
\thanks{$^{*}${\tt\small federico.moro@itia.cnr.it}}%
}

\begin{document}

\maketitle
\thispagestyle{empty}
\pagestyle{empty}

\begin{abstract}
This paper presents a concurrent whole-body control (cWBC) for human-exoskeleton systems that are tightly coupled at a Cartesian level (e.g., feet, hands, torso). The exoskeleton generates joint torques that i) cancel the effects of gravity on the coupled system, ii) perform a primary task (e.g., maintaining the balance of the system), and iii) exploit the kinematic redundancy of the system to amplify the forces exerted by the human operator. The coupled dynamic system is demonstrated to be passive, as its overall energy always goes dissipated until a minimum is reached. The proposed method is designed specifically to control exoskeletons for power augmentation worn by healthy operators in applications such as manufacturing, as it allows to increase the worker's capabilities, therefore reducing the risk of injuries.
\end{abstract}

\section{Introduction}
In recent years there has been a growing research and commercial interest in exoskeletons, and this is witnessed by the many new prototypes that have been presented both by research institutions [1]--[5] and private companies [6]--[9]. A typical classification distinguishes between exoskeletons for assistance and/or rehabilitation of disabled people, and exoskeleton for power augmentation of healthy people. The latter are envisioned to increase the workers' capabilities in scenarios like lifting and handling heavy objects (e.g., in manufacturing), and to reduce their risk to get injured.

Compared to medical exoskeletons, power augmentation exoskeletons are expected to demonstrate higher performances (i.e., to produce higher forces at higher velocities), and to guarantee transparency to the human operator. Moreover, the control action generated by the exoskeleton always has to comply with the control action generated by the operator, that will likely lead the movement. These requirements have to steer both the mechanical design of an exoskeleton, and its control.

This paper proposes a concurrent whole-body control (cWBC) formulation, that exploits the kinematic redundancy of a human-exoskeleton system that is coupled at a Cartesian level (e.g., feet, hands, torso) to perform multiple tasks. Whole-body control is a well-known control paradigm [10]--[14], with typical application to highly redundant legged robots (e.g., humanoids, quadrupeds). To the best of the authors' knowledge, this is the first time that whole-body control is applied to a human-exoskeleton system.

Section II presents the concurrent whole-body control (cWBC) formulation. The presence of a second, concurrent control action by the human operator makes the evaluation of the system stability more challenging. In Section III a demonstration of passivity is provided, under the assumption that the control action of the human is stable. It will be shown that all the energy of the dynamic system goes dissipated until a minimum is reached. The cWBC system is experimentally validated in simulation, and the results are analyzed in Section IV.

\section{The Concurrent Whole-Body Control (cWBC) Formulation}
A human-exoskeleton system, rigidly coupled at a Cartesian level (see Figure 1) can be modeled as follows:
\begin{figure}[!t]
\centering
\begin{subfigure}{0.23\textwidth}
\centering
\includegraphics[height=7cm]{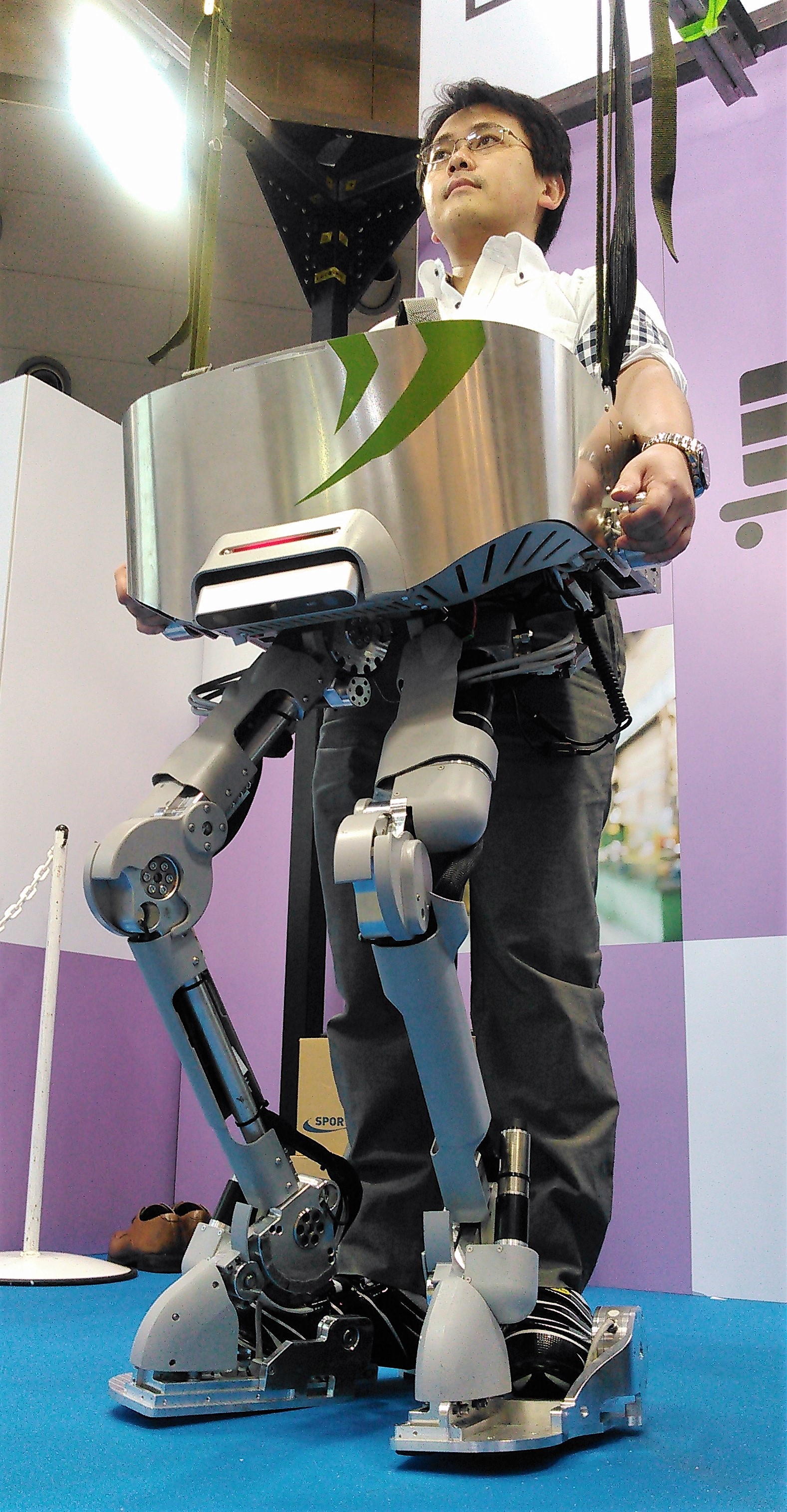}
\caption{}\label{fig:panasonic}
\end{subfigure}
\begin{subfigure}{0.23\textwidth}
\centering
\includegraphics[height=7cm]{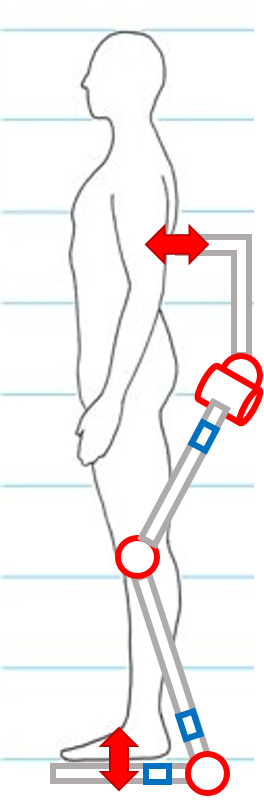}
\caption{}\label{fig:struct}
\end{subfigure}
\caption{(a) The MS-04 exoskeleton - \copyright 2016 Activelink Co., Ltd. - is a notable example of power augmentation exoskeleton that is coupled with the human operator at a Cartesian level, and can therefore be controlled using the concurrent whole-body control (cWBC) proposed in this paper, (b) a schema of lower-body exoskeleton with Cartesian coupling at the feet and torso}\label{fig:error}
\end{figure}
\begin{equation}
\begin{cases}
M_x \left( q_x \right) \ddot q_x + C_x \left( q_x, \dot q_x \right) + h_x \left( q_x \right) = \tau_x + J_{x}^T \left( q_x \right) f_{ox}\\
M_o \left( q_o \right) \ddot q_o + C_o \left( q_o, \dot q_o \right) + h_o \left( q_o \right) = \tau_o - J_{o}^T \left( q_o \right) f_{ox}
\end{cases}
\end{equation}
\noindent where subscripts $_x$ and $_o$ refer to exoskeleton and human operator quantities, respectively. $M(q)$, $C(q, \dot q)$, and $h(q)$ are the inertia matrix, the Coriolis and centripedal term, and the gravity vector, $q$, $\dot q$, and $\ddot q$ are joint positions, velocities, and accelerations, and $\tau$ are joint torques. $J_x(q_x)$ is the Jacobian that relates exoskeleton joint velocities and Cartesian velocities of the points of contact between exoskeleton and human operator. $J_o(q_o)$ is the Jacobian that relates human joint velocities and Cartesian velocities of the points of contact between exoskeleton and human operator. $f_{ox}$ is the human-exoskeleton interaction force. The exoskeleton friction forces $F_x \left( q, \dot q \right)$ are neglected, as they can always be estimated and compensated.

The following kinematic constraints also have to be imposed to guarantee a rigid coupling:
\begin{subequations}
\begin{align}
&x_o (q_o) = x_x (q_x)\\
&q_o = x_o^{-1} \left( x_x (q_x) \right)
\end{align}
\end{subequations}
\begin{subequations}
\begin{align}
&J_o \dot q_o = J_x \dot q_x\\
&\dot q_o = J_o^{\#} J_x \dot q_x
\end{align}
\end{subequations}
\begin{subequations}
\begin{align}
&\dot J_o \dot q_o + J_o \ddot q_o = \dot J_x \dot q_x + J_x \ddot q_x\\
&\ddot q_o = J_o^{\#} \left( J_x \ddot q_x + \dot J_x \dot q_x - \dot J_o \dot q_o \right) \nonumber\\
&\ddot q_o = J_o^{\#} \left( J_x \ddot q_x + \cancel{\dot J_x \dot q_x} - \cancel{\dot J_o J_o^{\#} J_x \dot q_x} \right) \nonumber\\
&\ddot q_o = J_o^{\#} J_x \ddot q_x
\end{align}
\end{subequations}
These constraints ensure that the points of contact between the human operator and the exoskeleton always have the same Cartesian position (Eq. 2a), velocity (Eq. 3a), and acceleration (Eq. 4a). Notice that, in the typical case of a redundant kinematic chain, the operator's joint positions $q_o$ (Eq. 2b), velocities $\dot q_o$ (Eq. 3b), and accelerations $\ddot q_o$ (Eq. 4b) cannot be derived uniquely from the Cartesian position $x_o$, velocity $\dot x_o$, and acceleration $\ddot x_o$. A reasonable inverse kinematic solutions $x_o^{-1}$, and a minimal norm pseudoinverse $J_o^{\#}=J_o^T \left( J_o J_o^T \right)^{-1}$ are used to obtain an approximated estimation, which is expected to be very close to the actual values in the case of e.g., the legs, where the limited range of motion of some joints reduces a lot the variability of the valid solutions. The higher order terms can be canceled for simplicity, as their effect at slow velocities is negligible.

Combining the two equations of the system in (1), it becomes:
\begin{align}
\begin{split}
M_x \ddot q_x + \cancel{C_x} + h_x - \tau_x &+\\
+ J^T_x J^{\#T}_o \left( M_o \ddot q_o + \cancel{C_o} + h_o - \tau_o \right) &= 0
\end{split}
\end{align}
As long as the dynamics is slow enough, the contribution of the Coriolis and centripedal term is negligible, and can therefore be omitted with no significant effects on the proposed control. For an increased accuracy these forces can always be estimated and compensated. Using Eq. 4b in Eq. 5, it becomes:
\begin{align}
\begin{split}
M_x \ddot q_x + h_x - \tau_x &+\\
+ J^T_x J^{\#T}_o \left( M_o J_o^{\#} J_x \ddot q_x + h_o - \tau_o \right) &= 0
\end{split}
\end{align}
$M_o(q_o)$, $h_o(q_o)$, $J_o(q_o)$ can become $M_o(q_x)$, $h_o(q_x)$, $J_o(q_x)$ using the transformation in Eq. 2b.

Rearranging Eq. 6, it becomes:
\begin{align}
\begin{split}
\left( M_x + J^T_x J^{\#T}_o M_o J_o^{\#} J_x \right) \ddot q_x + h_x + J^T_x J^{\#T}_o h_o =\\
= \tau_x + J^T_x J^{\#T}_o \tau_o
\end{split}
\end{align}
A control action $\tau_x$ has to be designed. First, a model-based feed-forward gravity compensation (of both exoskeleton and human) can be implemented: with ${}^1\tau_x = h_x + J^T_x J^{\#T}_o J^T_{co} m_o g$, Eq. 7 becomes:
\begin{align}
\begin{split}
\left( M_x + J^T_x J^{\#T}_o M_o J_o^{\#} J_x \right) \ddot q_x = J^T_x J^{\#T}_o \tau_o
\end{split}
\end{align}
Notice that to compensate the effects of gravity on the operator, an estimation of the human model is required. In particular, $J_{co}$, the Jacobian of the human Center of Mass (CoM), has to be estimated ($m_o$ can be easily measured).

A stable control action should be closed-loop. A task $1$ can be identified, and a force $f_1$ that minimizes the error on the task can be applied as following: ${}^2\tau_x = {}^1\tau_x + J^T_1 f_1$. With this torque applied by the exoskeleton, Eq. 7 becomes:
\begin{align}
\begin{split}
\left( M_x + J^T_x J^{\#T}_o M_o J_o^{\#} J_x \right) \ddot q_x = J^T_1 f_1 + J^T_x J^{\#T}_o \tau_o
\end{split}
\end{align}
Considered the goal to augment the force of the operator, a third contribution has to be specified. This second task is defined in the nullspace of the primary task, and amplifies $\tau_o$ by a factor of $k_{FF}$: ${}^3\tau_x = {}^2\tau_x + (J_x N_1)^T k_{FF} J^{\#T}_o \tau_o$. Therefore, Eq. 7 becomes:
\begin{align}
\begin{split}
&\left( M_x + J^T_x J^{\#T}_o M_o J_o^{\#} J_x \right) \ddot q_x =\\
= &J^T_1 f_1 + \left( J^T_x + (J_x N_1)^T k_{FF} \right) J^{\#T}_o \tau_o
\end{split}
\end{align}
A realistic primary task is balance maintenance of the coupled human-exoskeleton system. It can be implemented as a centroidal impedance: $J^T_1 f_1 = J^T_c \left( k_{Pc} ( x_c^d - x_c ) - k_{Dc} \dot x_c \right)$, where $J_c$ is the Jacobian of the human-exoskeleton Center of Mass (CoM), and $x_c^d$ is a desired CoM position (typically the center of the convex hull defined by the feet, with no vertical position imposed). $x_c$ and $\dot x_c$ are the actual position and velocity of the CoM, respectively, and $k_{Pc}$ and $k_{Dc}$ are coefficients of stiffness and damping. Eq. 10 can be rewritten as:
\begin{align}
\begin{split}
\left( M_x + J^T_x J^{\#T}_o M_o J_o^{\#} J_x \right) \ddot q_x =\\
= J^T_c \left( k_{Pc} ( x_c^d - x_c ) - k_{Dc} \dot x_c \right) +\\
+ \left( J^T_x + (J_x N_c)^T k_{FF} \right) J^{\#T}_o \tau_o
\end{split}
\end{align}
Assuming that a stable control action is performed by the human operator, e.g., $\tau_o = J^T_o \left( k_{Po} (x^d - x) - k_{Do} \dot x \right)$, where $x^d$ is a desired end-effector position, $x$ and $\dot x$ are its actual position and velocity, and $k_{Po}$ and $k_{Do}$ are coefficients of stiffness and damping, it becomes:
\begin{align}
\begin{split}
\left( M_x + J^T_x J^{\#T}_o M_o J_o^{\#} J_x \right) \ddot q_x =\\
= J^T_c \left( k_{Pc} ( x_c^d - x_c ) - k_{Dc} \dot x_c \right) +\\
+ \left( J^T_x + (J_x N_c)^T k_{FF} \right) \left( k_{Po} (x^d - x) - k_{Do} \dot x \right)
\end{split}
\end{align}
\begin{figure}[!t]
\vspace{0.3cm}
\centering
\includegraphics[width=\linewidth]{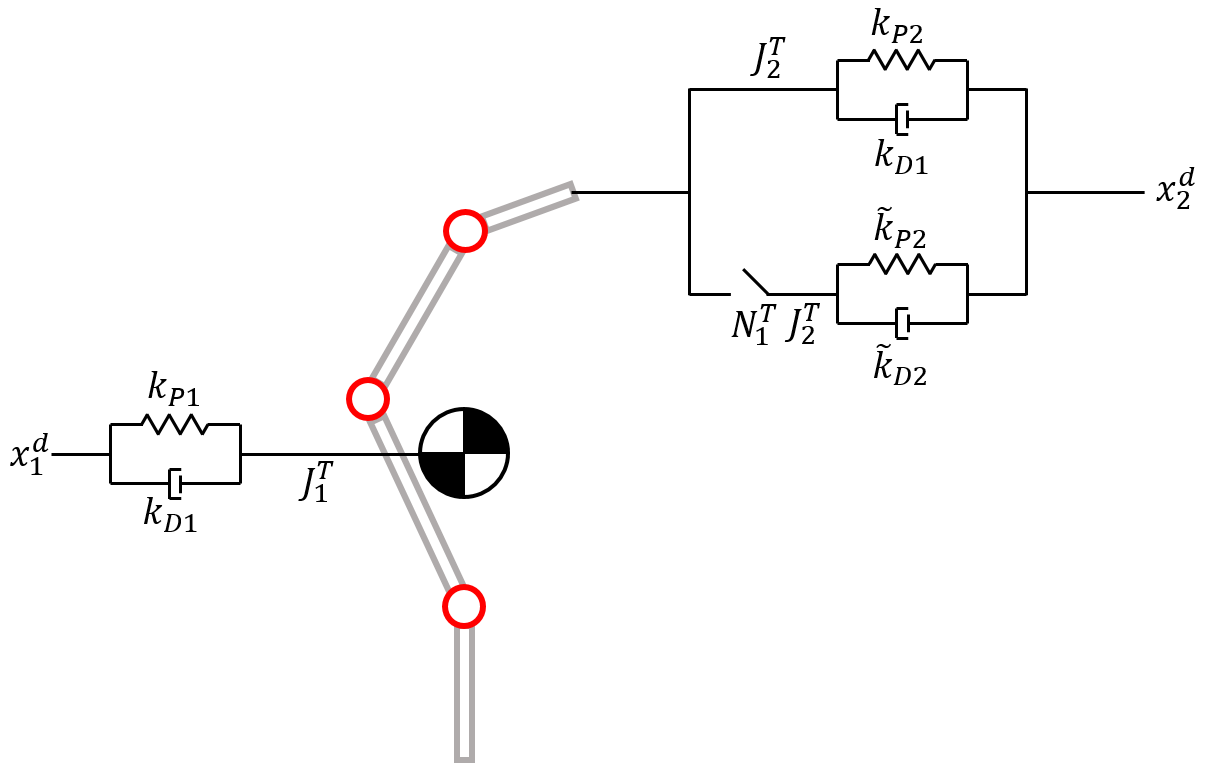}
\caption{A schematic representation of Eq. 15, that describes the dynamics of a coupled human-exoskeleton system that is subject to the torques generated by the proposed cWBC. A centroidal impedance is on the left, and is generated by the exoskeleton as a primary task. On the right is a Cartesian impedance, composed by the contribution of the human operator, that is potentially in conflict with the balancing task, and by the contribution of the exoskeleton, that is defined in the null space of the primary task, and amplifies the forces generated by the human operator.}\label{fig:model1}
\end{figure}
From the practical point of view, $\tau_o$ is not known, and cannot be measured directly. $f_{ox}$, instead, can be measured by a F/T sensor located at the point of contact between the human operator and the exoskeleton. The torque $\tau_x$ generated by the exoskeleton, therefore, becomes:
\begin{align}
\begin{split}
{}^3\tau_x &= {}^2\tau_x + (J_x N_1)^T k_{FF} J^{\#T}_o \left( M_o J_o^{\#} J_x \ddot q_x + h_o + J_{o}^T f_{ox} \right) =\\
&=h_x + J^T_x J^{\#T}_o J^T_{co} m_o g + J^T_1 f_1 +\\
&\hspace{0.4cm}+ (J_x N_1)^T k_{FF} J^{\#T}_o \left( M_o J_o^{\#} J_x \ddot q_x + h_o + J_{o}^T f_{ox} \right) =\\
&=h_x + J^T_x J^{\#T}_o J^T_{co} m_o g +\\
&\hspace{0.4cm}+ J^T_c \left( k_{Pc} ( x_c^d - x_c ) - k_{Dc} \dot x_c \right) +\\
&\hspace{0.4cm}+ (J_x N_c)^T k_{FF} J^{\#T}_o \left( M_o J_o^{\#} J_x \ddot q_x + h_o + J_{o}^T f_{ox} \right)
\end{split}
\end{align}
When the dynamics is slow enough, the inertial term in the feedback can be omitted. This simplification allows to avoid using $\ddot q_x$ as an input to the control. In this case, it becomes:
\begin{align}
\begin{split}
{}^3\tau_x =&h_x + J^T_x J^{\#T}_o J^T_{co} m_o g +\\
&+ J^T_c \left( k_{Pc} ( x_c^d - x_c ) - k_{Dc} \dot x_c \right) +\\
&+ (J_x N_1)^T k_{FF} J^{\#T}_o \left( h_o + J_{o}^T f_{ox} \right)
\end{split}
\end{align}
Notice that using the F/T sensor as described in Eqs. 13-14 works correctly only if the lifted object is in contact with the exoskeleton, and not with the human operator.

\section{Passivity of the Dynamic System}
A human-exoskeleton system tightly coupled at a Cartesian level can be modeled as in Eq. 1. A concurrent whole-body control (cWBC) was proposed in the previous section, and is described by Eq. 12.

Defining $M_{xo}=M_x + J^T_x J^{\#T}_o M_o J_o^{\#} J_x$, Eq. 12 can be written as:
\begin{align}
\begin{split}
M_{xo} \ddot q_x &= J^T_c \left( k_{Pc} ( x_c^d - x_c ) - k_{Dc} \dot x_c \right) +\\
&\hspace{0.4cm} + J^T_x \left( k_{Po} (x^d - x) - k_{Do} \dot x \right) +\\
&\hspace{0.4cm} + (J_x N_c)^T k_{FF} \left( k_{Po} (x^d - x) - k_{Do} \dot x \right)
\end{split}
\end{align}
The passivity of the dynamic system in Eq. 15, which is schematically represented in Figure 2, will be demonstrated by showing that the energy of the system is always dissipated until it reaches a minimum. Similar approaches have been used to demonstrate the passivity of whole-body control methods for redundant robots in [15]--[17].

To be more general, Eq. 15 can be rewritten as:
\begin{align}
\begin{split}
M \ddot q &= J^T_1 \left( k_{P1} ( x_1^d - x_1 ) - k_{D1} \dot x_1 \right) +\\
&\hspace{0.4cm} + J^T_2 \left( k_{P2} (x_2^d - x_2) - k_{D2} \dot x_2 \right) +\\
&\hspace{0.4cm} + (J_2 N_1)^T k_{FF} \left( k_{P2} (x_2^d - x_2) - k_{D2} \dot x_2 \right)
\end{split}
\end{align}
Defining $J^T_{2|1} = (J_2 N_1)^T$, $\tilde{k}_{P2} = k_{FF} k_{P2}$, and $\tilde{k}_{D2} = k_{FF} k_{D2}$, it becomes:
\begin{align}
\begin{split}
M \ddot q &= J^T_1 \left( k_{P1} ( x_1^d - x_1 ) - k_{D1} \dot x_1 \right) +\\
&\hspace{0.4cm} + J^T_2 \left( k_{P2} (x_2^d - x_2) - k_{D2} \dot x_2 \right) +\\
&\hspace{0.4cm} + J^T_{2|1} \left( \tilde{k}_{P2} (x_2^d - x_2) - \tilde{k}_{D2} \dot x_2 \right)
\end{split}
\end{align}
Considering the three control actions separately, it can be noticed that the first two contributions from Eq. 16 both have the form:
\begin{align}
\begin{split}
M \ddot q &= J^T_i \left( k_{Pi} ( x_i^d - x_i ) - k_{Di} \dot x_i \right) 
\end{split}
\end{align}
In Cartesian it becomes:
\begin{align}
\begin{split}
\Lambda_i \ddot x_i &= k_{Pi} ( x_i^d - x_i) - k_{Di} \dot x_i
\end{split}
\end{align}
where $\Lambda_i = J_i^{\#T} M J_i^{\#}$. The following variables are introduced:
\begin{align}
\begin{split}
&\varepsilon_i = x_i - x_i^d\\
&\dot \varepsilon_i = \dot x_i = J_i \dot q, \hspace{0.5cm} \dot q = J_i^{\#} \dot \varepsilon_i = J_i^{\#} \dot x_i
\end{split}
\end{align}
The kinetic energy of the system can be computed as:
\begin{align}
\begin{split}
E_c = \frac{1}{2} \dot x_i^T \Lambda_i \dot x_i
\end{split}
\end{align}
While the potential energy is:
\begin{align}
\begin{split}
U = \frac{1}{2} \varepsilon_i^T k_{Pi} \varepsilon_i
\end{split}
\end{align}
The total energy (Hamiltonian) is taken as storage function:
\begin{align}
\begin{split}
S &= E_c + U =\\
&= \frac{1}{2} \dot x_i^T \Lambda_i \dot x_i + \frac{1}{2} \varepsilon_i^T k_{Pi} \varepsilon_i
\end{split}
\end{align}
Since $\Lambda_i$, and $k_{Pi}$, are positive definite matrices, $S$ is also positive definite. Its derivative $\dot S$ is:
\begin{align}
\begin{split}
\dot S &= \dot x_i^T \Lambda_i \ddot x_i + \varepsilon_i^T k_{Pi} \dot \varepsilon_i
\end{split}
\end{align}
Using Eq. 19 in Eq. 24, it becomes:
\begin{align}
\begin{split}
\dot S &= - \dot x_i^T \left( k_{Pi} \varepsilon_i + k_{Di} \dot x_i \right) + \dot \varepsilon_i^T k_{Pi} \varepsilon_i
\end{split}
\end{align}
\begin{figure}[!t]
\centering
\includegraphics[width=0.95\linewidth]{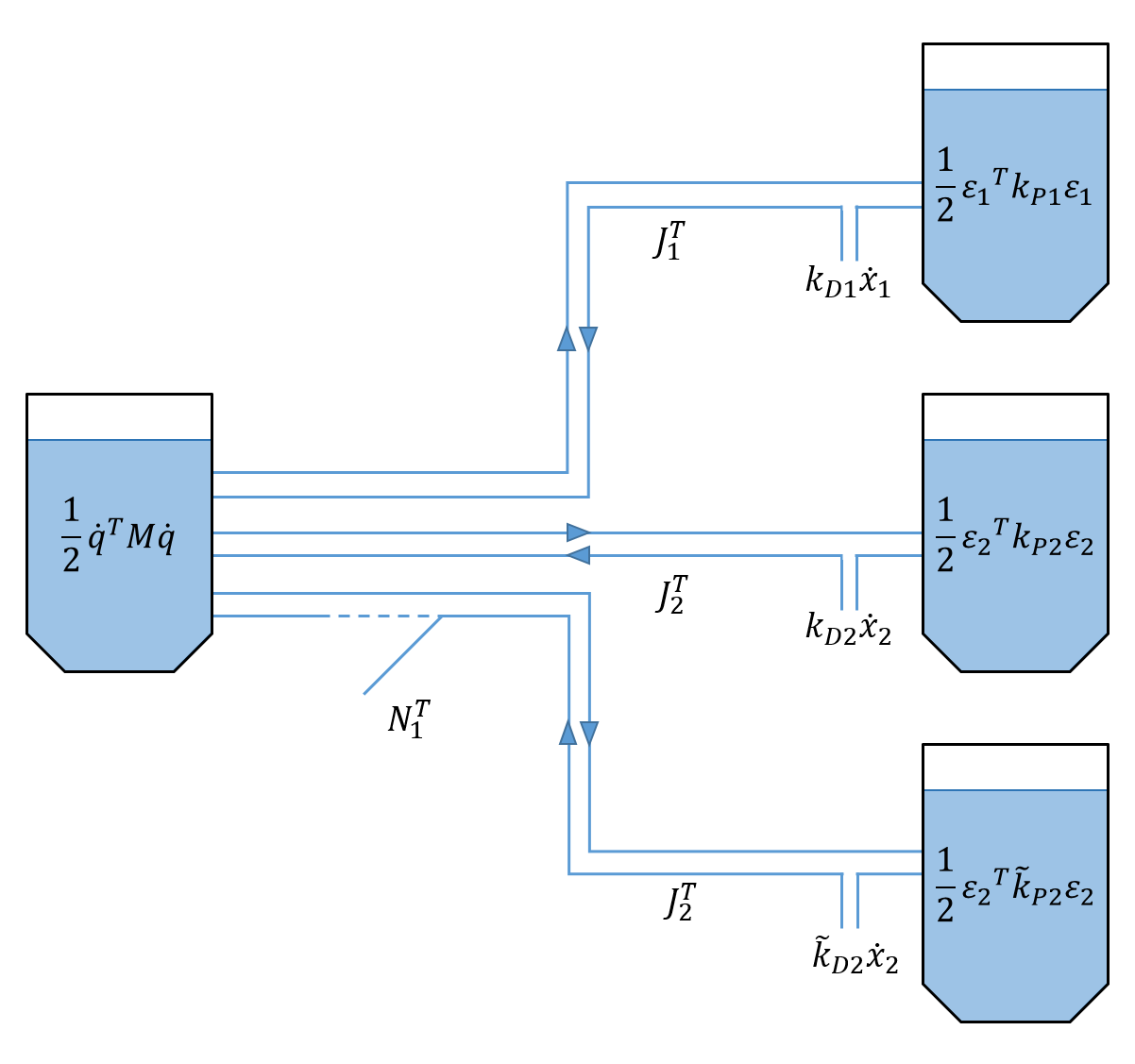}
\caption{A visual representation of the energy flow of the system analized. It is composed by energy storing elements, power-preserving connections, a switch, and dissipative elements. This means that the system is passive, as it dissipates energy until a minimum is reached.}\label{fig:model2}
\end{figure}
Using Eq. 20, it can be rewritten as:
\begin{align}
\begin{split}
\dot S &= - \cancel{\dot \varepsilon_i^T k_{Pi} \varepsilon_i} - \dot x_i^T k_{Di} \dot x_i + \cancel{\dot \varepsilon_i^T k_{Pi} \varepsilon_i}
\end{split}
\end{align}
Being $k_{Di}$ positive definite, $\dot S$ is always negative definite. This is an expected result, as the kinetic energy storing and the potential energy storing are atomic energy storing elements, and are passive and irreducible. The Jacobians represent internal power-preserving interconnections, and the damping is an energy dissipation element.

Moreover, Eq. 19 has the typical form of the free vibration with damping, which is well known to be stable as it can be written in state space equations, and its eigenvalues always have a real part that is negative.

The third control action in Eq. 16 is slightly different, as it is defined in the null space of the first one, and has therefore the following form:
\begin{align}
\begin{split}
M \ddot q &= J_{i|j}^T \left( \tilde{k}_{Pi} ( x_i^d - x_i ) - \tilde{k}_{Di} \dot x_i \right) 
\end{split}
\end{align}
where $j$ is a task with higher priority than task $i$. In Cartesian it becomes:
\begin{align}
\begin{split}
\Lambda_{i|j} \ddot x_i &= \tilde{k}_{Pi} ( x_i^d - x_i) - \tilde{k}_{Di} \dot x_i
\end{split}
\end{align}
where $\Lambda_{i|j} = J_{i|j}^{\#T} M J_{i|j}^{\#}$. As it was done before, the following variables are introduced:
\begin{align}
\begin{split}
&\varepsilon_i = x_i - x_i^d\\
&\dot \varepsilon_i = \dot x_i = J_i \dot q, \hspace{0.5cm} \dot q = J_{i|j}^{\#} \dot \varepsilon_i = J_{i|j}^{\#} \dot x_i\\
\end{split}
\end{align}
\begin{figure}[!t]
\centering
\begin{subfigure}{0.5\textwidth}
\centering
\includegraphics[width=0.985\linewidth]{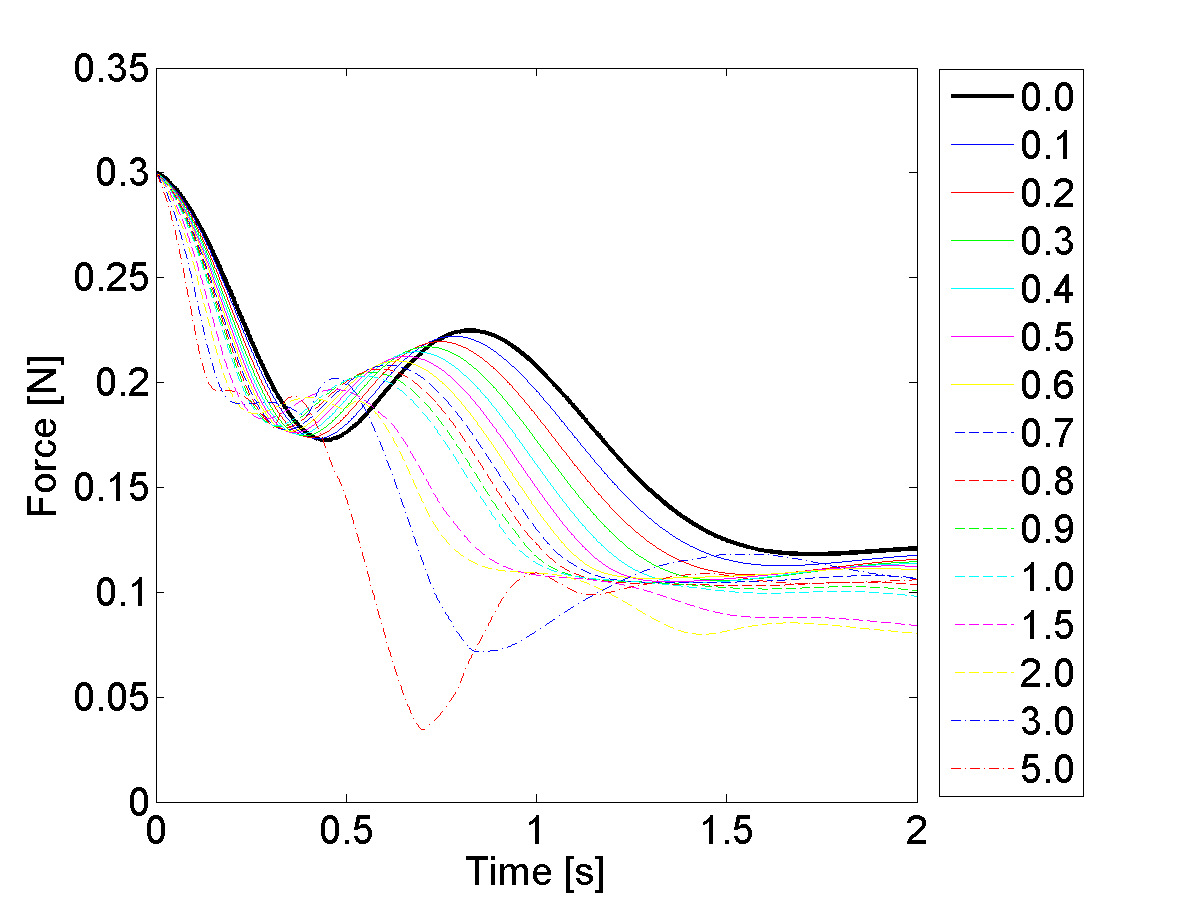}
\caption{}\label{fig:force_x}
\end{subfigure}
\begin{subfigure}{0.5\textwidth}
\centering
\includegraphics[width=0.985\linewidth]{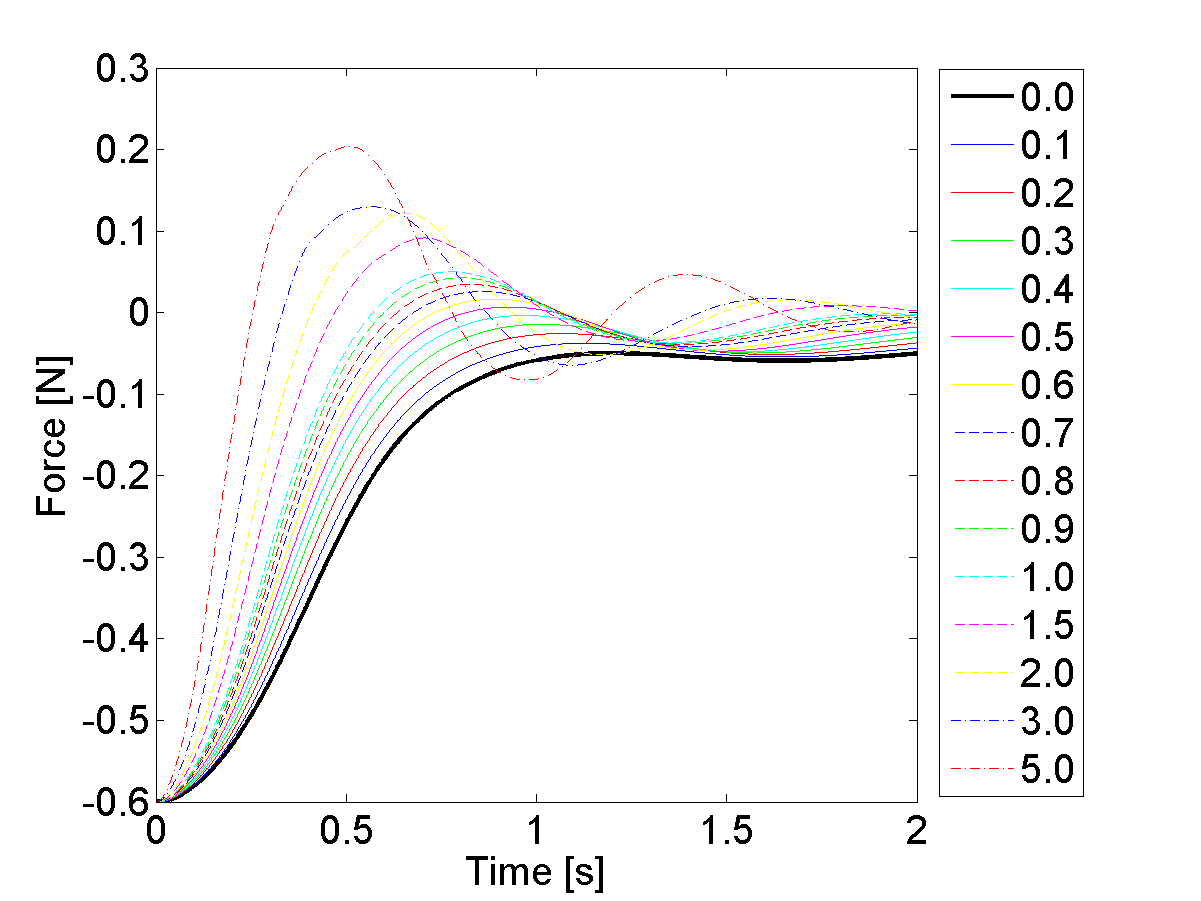}
\caption{}\label{fig:force_z}
\end{subfigure}
\begin{subfigure}{0.5\textwidth}
\centering
\includegraphics[width=0.985\linewidth]{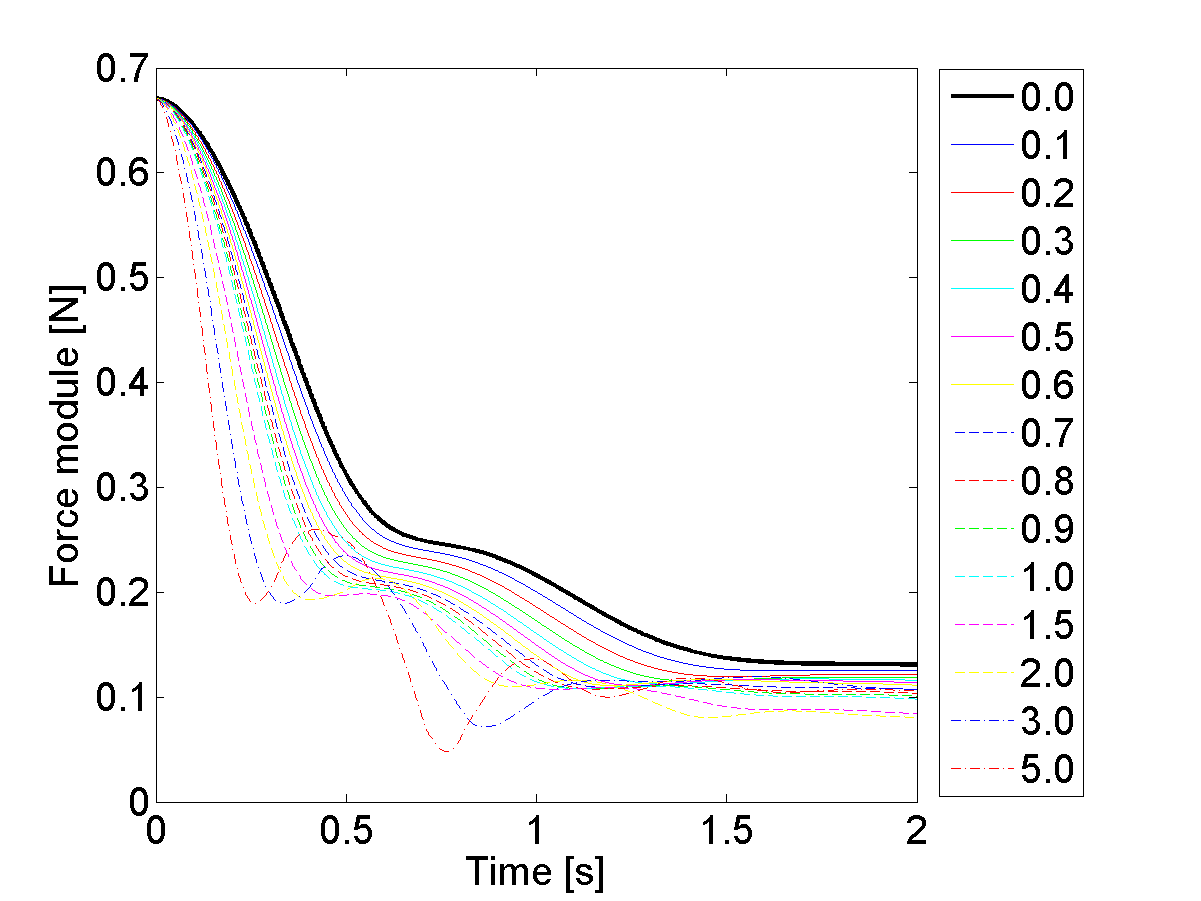}
\caption{}\label{fig:force_module}
\end{subfigure}
\caption{The measured interaction force between human and exoskeleton (a) in the horizontal direction, (b) in the vertical direction, and (c) its module, at different force augmentation factors.}\label{fig:force}
\end{figure}
The total energy (Hamiltonian) is taken as storage function:
\begin{align}
\begin{split}
S = \frac{1}{2} \dot x_i^T \Lambda_{i|j} \dot x_i + \frac{1}{2} \varepsilon_i^T \tilde{k}_{Pi} \varepsilon_i
\end{split}
\end{align}
Since $\Lambda_{i|j}$ is a positive semidefinite matrix, and $\tilde{k}_{Pi}$ is a positive definite matrix, $S$ is also positive semidefinite. Its derivative $\dot S$ is:
\begin{align}
\begin{split}
\dot S &= \dot x_i^T \Lambda_{i|j} \ddot x_i + \varepsilon_i^T \tilde{k}_{Pi} \dot \varepsilon_i
\end{split}
\end{align}
Using Eq. 28 in Eq. 31, it becomes:
\begin{align}
\begin{split}
\dot S &= - \dot x_i^T \left( \tilde{k}_{Pi} \varepsilon_i + \tilde{k}_{Di} \dot x_i \right) + \dot \varepsilon_i^T \tilde{k}_{Pi} \varepsilon_i
\end{split}
\end{align}
Eq. 29 is then used to obtain:
\begin{align}
\begin{split}
\dot S &= - \cancel{\dot \varepsilon_i^T \tilde{k}_{Pi} \varepsilon_i} - \dot x_i^T \tilde{k}_{Di} \dot x_i + \cancel{\dot \varepsilon_i^T \tilde{k}_{Pi} \varepsilon_i}
\end{split}
\end{align}
Being $\tilde{k}_{Di}$ positive definite, we know that $\dot S$ is negative definite. Differently from the previous case, this control action is defined in the null space of the first control action. If the redundancy of the system is not sufficient to perform both tasks, the secondary task is not performed, i.e., no energy is exchanged. Otherwise, if both tasks can be performed simultaneously, the behavior of this control action is the same as in the previous cases.

In summary, the system analyzed has four storing elements (one kinetic energy tank, and three potential energy tanks), which are connected by Jacobians, which preserve power. The only exception is on the third control action, which is defined in the null space of the first one. In this case the null space acts like a switch: it can let the energy flow, or stop it. In any case, it will never generate nor dissipate energy. The dissipation, instead, comes from the damping elements. Figure 3 provides a visual representation of the energy flow as it is described.

\section{Experimental Validation}
The concurrent whole-body control (cWBC) that is proposed in Section II, and whose passivity is demonstrated in Section III, was validated in Matlab on a simple, ideal model.

A planar hyper-redundant (5 degrees of freedom) manipulator represents the exoskeleton. The human operator is rigidly attached to its end-effector, and is modeled as a point mass. The exoskeleton generates a torque that i) compensates the effects of gravity on both human and exoskeleton, ii) minimizes the error on the horizontal COM position (i.e., implements a centroidal impedance), and iii) amplifies the force generated by the operator, which minimizes the error on the Cartesian position of the end-effector (i.e., produces a Cartesian impedance). This behavior is described by Eq.12.

Since the exoskeleton is redundant with respect to the two tasks that are imposed (1 degree of freedom for the horizontal position of the COM, 2 degrees of freedom for the Cartesian position of the end-effector, no orientation imposed), an additional joint momentum \cite{Moro4} damping is used to avoid drifting. The control frequency is set at 1 kHz.

At initial time both COM horizontal position and end-effector position have a non-zero error with respect to a fixed reference. The same experiment with identical initial conditions is repeated with different force augmentation factors.
\begin{figure}[!t]
\centering
\includegraphics[width=0.97\linewidth]{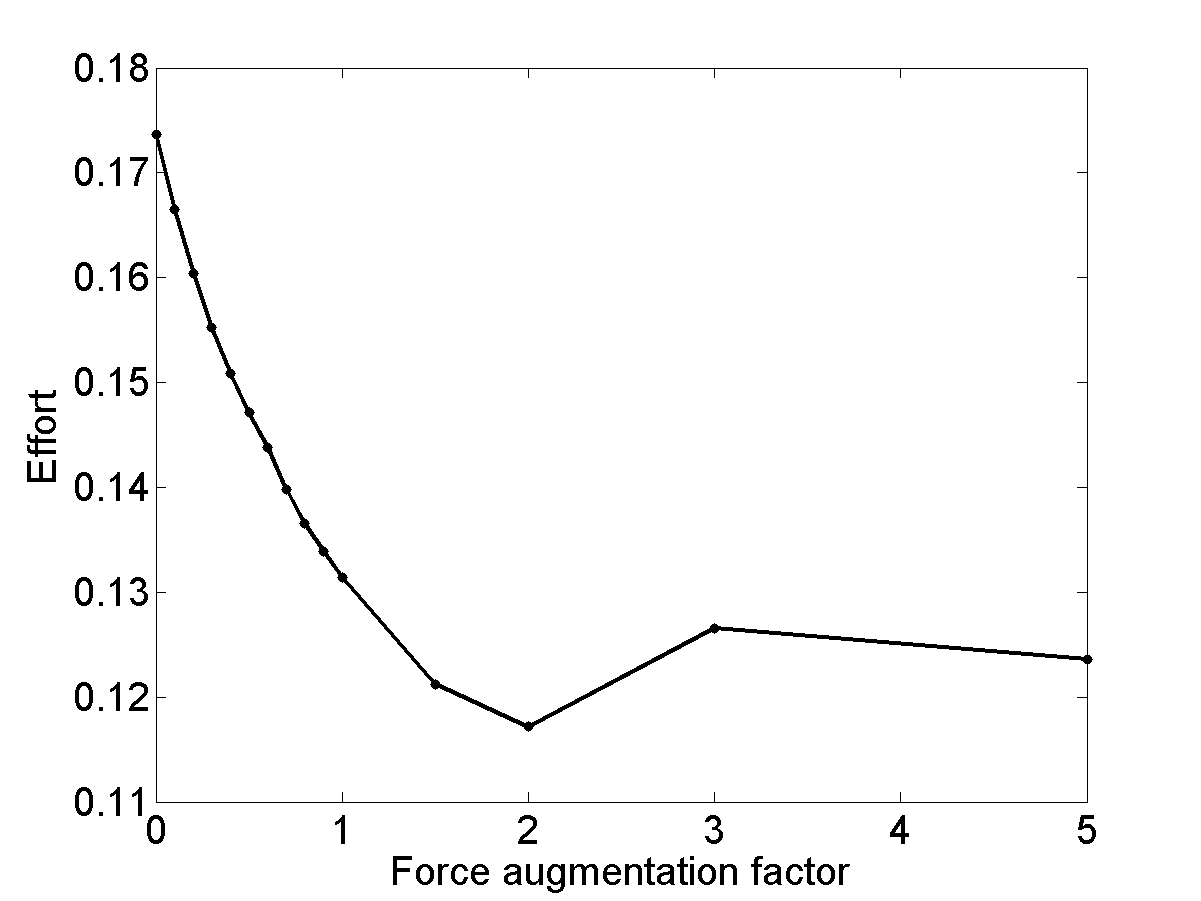}
\caption{The effort of the human operator at different force augmentation factors.}\label{fig:effort}
\end{figure}
\begin{figure}[!t]
\centering
\begin{subfigure}{0.5\textwidth}
\centering
\includegraphics[width=0.97\linewidth]{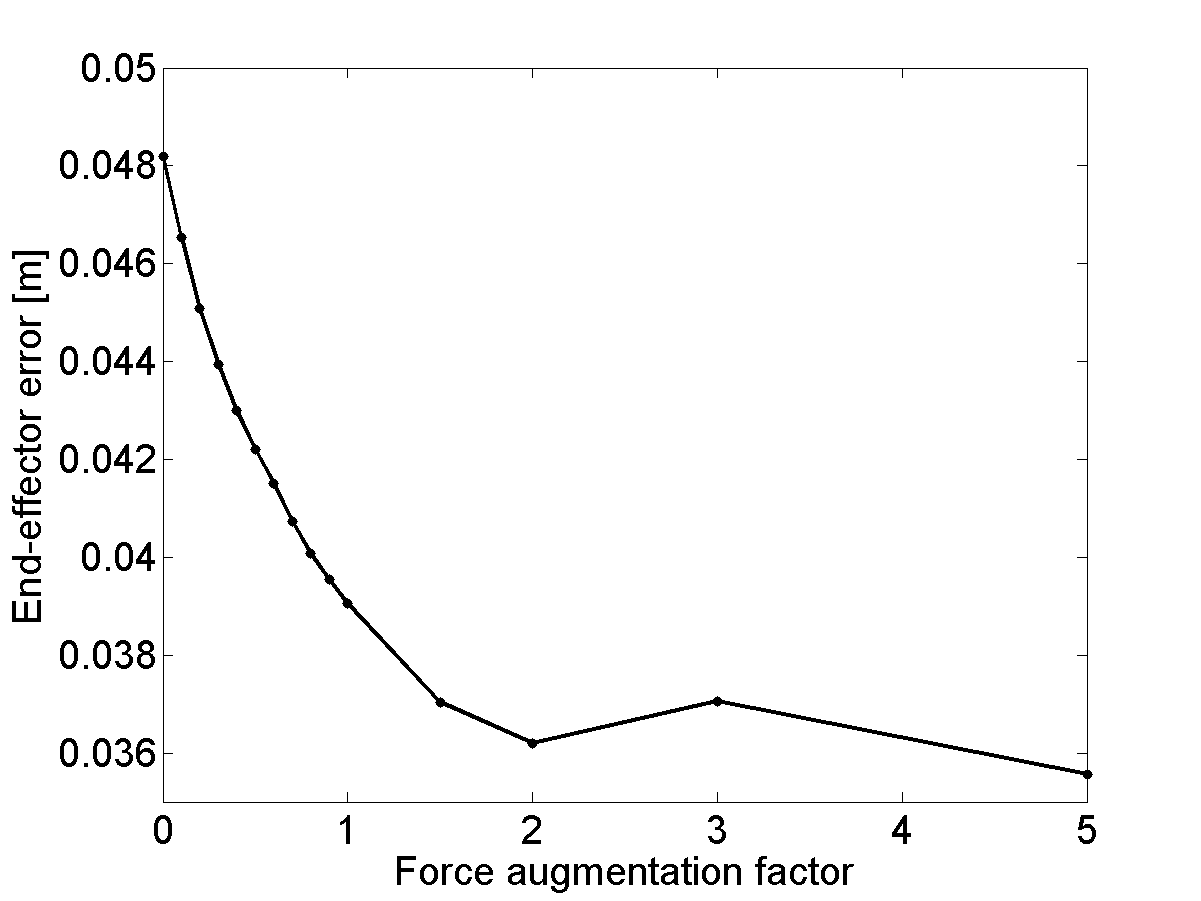}
\caption{}\label{fig:eeff_err}
\end{subfigure}
\begin{subfigure}{0.5\textwidth}
\centering
\includegraphics[width=0.97\linewidth]{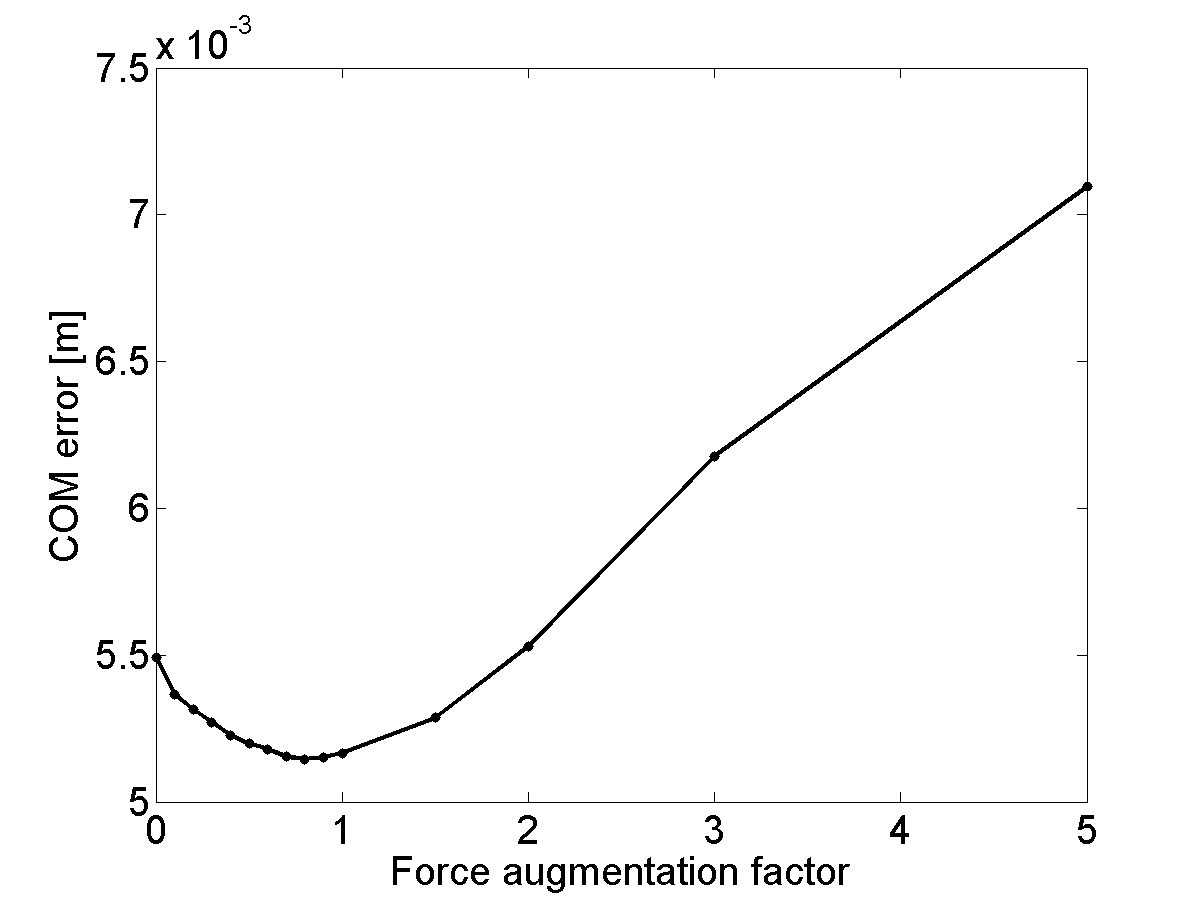}
\caption{}\label{fig:com_err}
\end{subfigure}
\caption{(a) Error on the end-effector position, and (b) error on the COM horizontal position, at different force augmentation factors.}\label{fig:error}
\end{figure}

Figures 5a and 5b show the measured interaction force between the exoskeleton and the human operator in the horizontal and in the vertical direction, respectively. The module of the interaction force is also reported in Figure 5c. It can be noticed that the interaction force reduces as the force augmentation factor grows. In all cases the control is stable and reaches an equilibrium. When $k_{FF}$ is higher than 1.5/2.0 some oscillations appear: this is very likely related to the control bandwidth.

From these data it is possible to derive the effort of the operator, computed as the squared of the force he generates to perform the reaching task. Figure 5 shows that the effort reduces as the augmentation factor grows (with $k_{FF}=2$ the effort is about 35\% less than with no force augmentation), as long as no oscillation appear. The operator is required to exert less force to perform the same task. Moreover, the exoskeleton is also compensating the gravitational effects on the operator, therefore reducing his effort even more if compared to free motion with no exoskeleton. 

It can also be interesting to notice that, as long as no oscillation appears, the error on the end-effector position at steady state (Figure 6a) significantly reduces, and the error on the COM horizontal position (Figure 6b) remains in the order of the millimeters.

\section{Final Remarks}
This paper presents a novel concurrent whole-body control (cWBC) for human-exoskeleton systems that are rigidly coupled at a Cartesian level. The cWBC exploits the redundancy of the coupled system, and allows the exoskeleton to perform multiple tasks: i) to compensate the effects of gravity on the coupled system, ii) to perform a primary task of balance maintenance (by means of a centroidal impedance), and iii) to augment the forces generated by the human operator, in the null space of the primary task.

The coupled dynamic system subject to the cWBC forces generated by the exoskeleton, and to a stable control action produced by the operator is demonstrated to be passive, as its overall energy always goes dissipated. This is an expected result as the system is composed by four storing elements (one kinetic energy tank, and three potential energy tanks), which are connected by Jacobians, which preserve power. The null space projection acts like a switch that either lets the energy flow or stops it, but does never produce nor dissipate energy. The damping terms, instead, dissipate energy until an equilibrium is reached at minimal energy conditions.

The cWBC is validated in simulation with a simple, ideal model. It is shown that the effort of the human operator significantly reduces with the force augmentation factor growing, and as long as no oscillation appears.

The proposed methodology allows to augment the operator's capabilities, although leaving him the movement lead. This is realistically expected to increase the transparency of the exoskeleton to the operator. Moreover, it does not require any intention detector (e.g., EMG measures, gait phase identification) but F/T sensors placed at the point of contact between the exoskeleton and the human operator.

Future work includes testing the system with a real power augmentation exoskeleton worn by a human operator. Using the cWBC on a real system requires to partly estimate the model of the human: this will necessarily introduce an error, but it is not expected to affect the overall behavior of the system.

\addtolength{\textheight}{-12cm}   

%

\end{document}